\documentclass[11pt]{article}

\usepackage[final]{acl}

\usepackage{times}
\usepackage{latexsym}

\usepackage[T1]{fontenc}

\usepackage[utf8]{inputenc}

\usepackage{microtype}

\usepackage{inconsolata}

\usepackage{graphicx}

\usepackage{color, soul}
\usepackage{makecell}
\usepackage{booktabs}
\usepackage{comment}
\usepackage{cleveref}

\title{Scaling E-Commerce Attribute Extraction with Parallel Decoding}

\author{
  \textbf{Nikhita Vedula} \quad
  \textbf{Dushyanta Dhyani} \quad
  \textbf{Bryan Wang} \quad
  \textbf{Shervin Malmasi} \\
  \vspace{0.1pt} \\
  Amazon.com, Inc. \\
  \small{\texttt{\{veduln, dhyanidd, brywan, malmasi\}@amazon.com}}
}

\begin{document}
\maketitle
\begin{abstract}

Customers rely on specific product attributes to compare products and make purchasing decisions, but e-commerce catalogs are messy and unstructured, making it difficult to identify which attributes matter most and extract them at scale. Standard Attribute Value Extraction (AVE) systems treat all attributes equally, producing large, inconsistent attribute sets that do not reflect the factors consumers use to differentiate products. We introduce a two-stage LLM pipeline that first discovers a compact, ranked schema of purchase-discriminative attributes for each product category, then extracts their values from catalog text using a fine-tuned compact LLM (Qwen3-4B) with Hyper-Parallel Decoding (HPD). %
This pipeline achieves 85\% extraction accuracy, on par with the foundational LLM it was distilled from, while reducing inference costs by 92\% over foundational LLMs, enabling production-scale use for product discovery and catalog enrichment. The resulting category-level structured representations effectively constitute automatically constructed product knowledge bases, providing consistent, comparable attributes across varied product categories that can ground downstream knowledge-intensive applications.

\end{abstract}

\section{Introduction}

Attribute Value Extraction is a core capability for e-commerce systems. Structured product attributes power search, filtering, comparison, recommendation, and product quality workflows \cite{yang2022mave,brinkmann2023extractgpt}. However, e-commerce catalogs are highly messy as product information is spread across titles, descriptions, bullets, and semi-structured fields; sellers use inconsistent terminology; and values are often missing, implicit, duplicated, or expressed in non-standard forms \cite{khandelwal2023large,yang2022mave}.
Recent work has increasingly explored generative approaches to AVE, where encoder-decoder models and LLMs extract attribute values directly from product text \cite{blume2023generative,shinzato2023unified,khandelwal2023large}.
These methods handle heterogeneous inputs and flexible output schemas, but deploying them at e-commerce scale introduces two major challenges. First, defining high-quality schemas manually across thousands of product categories is expensive and difficult to maintain \cite{huang2025attributeforge,xu2023towards}. Second, extracting values over massive catalogs is computationally costly, especially when products are added daily and schemas evolve with downstream needs \cite{chen2023named,yang2024eave,zhang2024stronger}.

We focus on extracting customer-relevant, purchase-discriminative attributes for each product category. Unlike standard AVE formulations that target broad attribute coverage or extract values for predefined attributes \cite{yang2022mave,xu2023towards}, our goal is to identify the attributes most useful for differentiating products within a category, such as wattage for blenders, capacity for storage devices, or noise cancellation for headphones, and ensure all products in the same category share the same schema, making their structured representations directly comparable. Therefore, our pipeline automatically induces a domain ontology (category-level schemas) and populates it at scale, constructing a structured product knowledge base without manual schema engineering.

We propose a scalable two-stage LLM pipeline. First, we automatically discover category-specific schemas from representative catalog examples using a large, foundational LLM (Claude), then standardize semantically similar attributes across categories.
In the second stage, we extract values for the discovered schemas from product catalog text using a fine-tuned small language model with Hyper-Parallel Decoding (HPD). HPD exploits the conditional independence of attribute values to decode multiple values in parallel \cite{glavas2026breaking}, substantially improving throughput.

We evaluate the system on a large proprietary e-commerce catalog spanning thousands of categories. Our schema discovery stage produces attributes judged correct or relevant in 89.6\% of cases, and our fine-tuned value extraction model achieves approximately 85\% extraction accuracy, matching the foundational LLM teacher model despite a $>$100$\times$ reduction in model parameters. The proposed system reduces inference cost by 92\%, making repeated catalog-scale processing practical.
Our main contributions are:

    (i) A two-stage pipeline that automatically discovers compact, category-level attribute schemas and extracts their values at scale, requiring no manual schema engineering.
    
    (ii) A practical, large-scale use case combining knowledge distillation into a compact LLM (Qwen3-4B) with Hyper-Parallel Decoding, processing hundreds of millions of products across multiple marketplaces while achieving 92\% cost reduction over a foundational LLM baseline.
    
    (iii) A comprehensive evaluation covering schema quality, extraction accuracy, quantity understanding, determinism, and efficiency, with detailed cost and throughput analysis.

\section{Related Work}

\paragraph{Product attribute-value extraction.}
Attribute Value Extraction (AVE) for e-commerce recovers structured product attributes from noisy catalog content to support search, recommendation, comparison, and catalog enrichment \cite{yang2022mave,brinkmann2023extractgpt}. Prior work has formulated AVE as sequence tagging \cite{zheng2018opentag} or question answering over product context \cite{wang2020learning}. These approaches improve extraction for predefined attributes, but typically assume available schemas and can become costly across many attributes and categories.

\paragraph{Generative AVE.}
Recent work has explored generative AVE, using encoder-decoder or LLM-based models to produce attribute values directly from product text \cite{shinzato2023unified,khandelwal2023large,brinkmann2023extractgpt}. These methods handle heterogeneous inputs and flexible output formats, but production-scale deployment still requires scalable schema construction and efficient extraction over large catalogs. Our work targets both challenges through automatic schema discovery and low-cost generative extraction.

\paragraph{Schema discovery and knowledge base construction.}
Open-world attribute mining and product schema modeling aim to discover or maintain product attributes with limited human supervision \cite{xu2023towards,huang2025attributeforge}. More broadly, automated knowledge base construction encompasses the extraction, integration, and maintenance of structured knowledge from unstructured sources \cite{weikum2021machine}, with recent work leveraging LLMs to reshape the classical pipeline of ontology engineering, knowledge extraction, and knowledge fusion \cite{bian2025llmkg}. In the e-commerce domain, \citet{hongwimol2026autopkg} present a multi-agent LLM framework that induces product types and attribute keys on demand and consolidates them into a globally consistent product knowledge graph, while \citet{peshevski2025agent} automate ontology creation and KG population from unstructured product descriptions without predefined schemas. Our work shares the goal of automated product KB construction but differs in emphasis: rather than maximizing broad attribute coverage, we seek compact, purchase-discriminative category-level schemas and prioritize extreme extraction efficiency through parallel decoding, enabling scaling to many millions of products.

\paragraph{Efficient and scalable extraction.}
Scalability is a central challenge for AVE because real-world products are associated with many attributes. Some QA-style approaches repeatedly process the same product context for different target attributes \cite{chen2023named}. Intra-prompt parallel decoding \cite{glavas2026intrapromptparalleldecodingcommoncontext} addresses this common-context bottleneck by stacking multiple questions into a single prompt and decoding their answers simultaneously through attention mask manipulation. Recent work improves efficiency by caching context representations and using lightweight attribute-context interactions \cite{yang2024eave}, while ``life-long'' AVE addresses the need to adapt to evolving products, categories, and attributes \cite{zhang2024stronger}. Our extraction stage is complementary: we specialize a small language model for structured value extraction and use Hyper-Parallel Decoding (HPD) to decode multiple conditionally independent attribute values from the same product context in parallel \cite{glavas2026breaking}.

\section{Scalable Schema-Guided Attribute Value Generation}

An e-commerce catalog contains billions of products, each with a unique set of attributes and values. Extracting these attribute-value pairs for each product in isolation can produce a large number of semantically similar but differently named attributes across the catalog, making them both unmanageable and difficult to use downstream. We propose a two-stage pipeline where we first identify a compact set of key purchase-discriminative attributes per product category, and subsequently extract values corresponding to each attribute for every product per category. Note that the choice of categorization granularity is important: a broad categorization with few categories will result in generic attributes, while being overly granular can result in an untractable attribute set size.

\begin{figure*}[t]
\centering
\includegraphics[width=\textwidth]{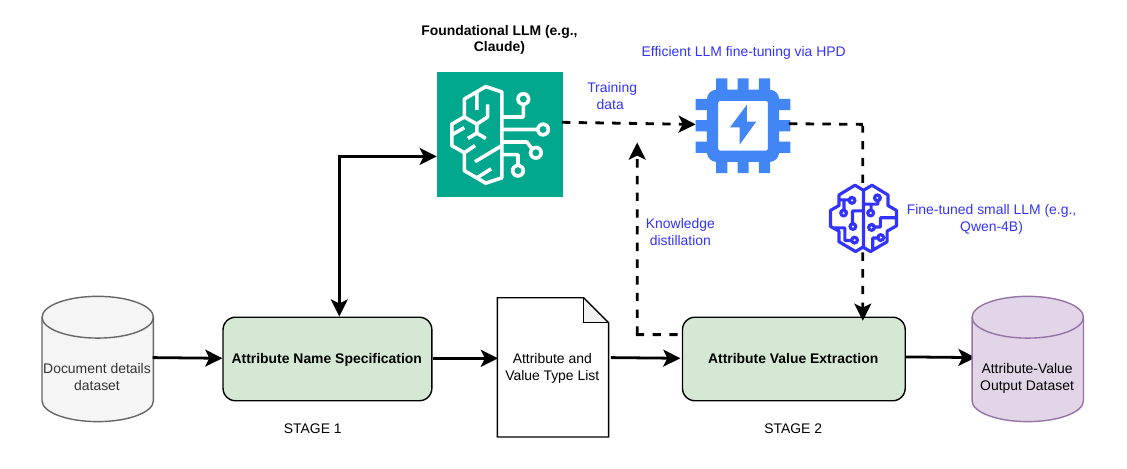}
\caption{Overview of our two-stage pipeline for scalable schema-guided attribute value generation. Stage~1 discovers a compact schema of $N$ purchase-discriminative attributes per category using a large foundational LLM. Stage~2 extracts attribute values from product catalog text using a fine-tuned compact LLM with Hyper-Parallel Decoding, generating all $N$ values simultaneously.}
\label{fig:architecture}
\end{figure*}

\subsection{Problem Statement}

Let $\mathcal{P}$ denote a large-scale e-commerce product catalog, categorized into $|\mathcal{T}|$ high-level categories. Each product $p \in \mathcal{P}$ has catalog context $\mathbf{x}_p$ comprising its title, description, and other unstructured product details, and is associated with a category $t(p) \in \mathcal{T}$.
We seek to produce, for each product $p$, a structured representation $\phi(p) \;=\; \bigl\{\,(a_i,\; v_i)\bigr\}_{i=1}^{N}$
where $\mathcal{A}_{t(p)} = \{a_1, \ldots, a_N\}$ is a schema of $N$ purchase-discriminative attributes defined at the category level, and each $v_i$ is the value of attribute $a_i$ extracted from $\mathbf{x}_p$ if available. This decomposes into two sub-problems:

\textbf{Attribute Schema Discovery:} For each category $t \in \mathcal{T}$, identify a schema $\mathcal{A}_t = \{a_1, \ldots, a_N\}$ of attributes that are most informative for consumer purchase decisions within that category.

\textbf{Attribute Value Generation:} For each product $p \in \mathcal{P}$, given the schema $\mathcal{A}_{t(p)}$ and the catalog context $\mathbf{x}_p$, extract the value $v_i$ of each attribute $a_i \in \mathcal{A}_{t(p)}$ from $\mathbf{x}_p$.
The key design constraint is \textit{category-level schema sharing}: all products belonging to the same category $t$ share the same attribute schema $\mathcal{A}_t$, ensuring that $\phi(p)$ and $\phi(p')$ are comparable for any two products $p, p'$ with $t(p) = t(p')$.

\subsection{Stage 1: Attribute Schema Discovery}

\paragraph{Category-Specific Attributes Specification}

For each category, we randomly select $k$ representative products that satisfy a minimum threshold of customer engagement metrics, ensuring a balanced selection of quality and diversity. We then use the Claude Sonnet 4 LLM to identify $M$ key attributes for the category from the title and description of each product. The prompts are tuned to also extract each attribute's description, data type (numerical, categorical, single-value, multi-valued, or free-form text), and a list of standardized values the attribute can take for the given category.

\paragraph{Global Attributes Specification}

While each product has its unique set of attributes (e.g., noise canceling ability for headphones or storage capacity for a hard disk), certain attributes are universally applicable across all products. We supplement the category-specific attributes with the following set of global attributes: \textit{total pack size, total weight, total volume, per product weight, per product volume.}

\paragraph{Attribute Name Standardization}

While defining a fixed set of $M$ attributes per category constrains the local schema, running Stage~1 across $P$ categories can produce $M \times P$ attributes (which can range from a few to tens of thousands) with semantic redundancy (e.g., \textit{Water Resistance, Waterproof rating, IP Water Rating}). We develop an LLM-based clustering approach to standardize these. We embed all attribute names using a Qwen3-8B embedding model and apply hierarchical agglomerative clustering to group semantically similar attributes. Each cluster is then assigned a canonical name via one of two methods: a) \textbf{Heuristic:} replacing all cluster members with the most frequent representative name, provided it satisfies a minimum similarity threshold; or b) \textbf{LLM-based:} using an LLM to generate a standardized name while preserving domain-specific context.   

\subsection{Stage 2: Parallel Attribute Value Generation}
\label{sec:stage2}

Once a standardized attribute set has been generated, the next step is to extract values for those attributes from all catalog products. A straightforward approach is to prompt a foundational LLM with the predefined attributes and product catalog information. While state-of-the-art LLMs can accomplish this with high accuracy, scaling to hundreds of millions of products in a cost-effective way poses a significant challenge. Even at an optimistic throughput of 100K products per hour, processing a complete catalog would take multiple weeks, which does not scale for a real-world e-commerce service.

Given a product context and a fixed schema of $N$ attributes, standard autoregressive decoding generates the output JSON sequentially, requiring approximately $\sum_{i=1}^{N} K_i$ decoding steps, where $K_i$ is the token length of the value for attribute $i$.

\paragraph{Hyper-Parallel Decoding (HPD): Efficiently Scaling Inference Throughput and Cost}

In schema-guided AVE, many attribute values are conditionally independent given the same product context. HPD \cite{glavas2026breaking} exploits this structure by decoding values for multiple attributes in parallel, reducing decoding steps from $\sum K_i$ to approximately $\max_i K_i$. Standard autoregressive decoding generates attribute-value pairs sequentially, attending to all previously generated values before producing the next. HPD observes that conditioned on the same product context $x_p$, the values of different attributes are conditionally independent: extracting a blender's wattage should not causally depend on having first extracted its color. This conditional independence allows HPD to extract values in parallel during decoding.

The mechanism works by constructing a JSON skeleton template with attribute keys and blank value fields, creating position ID ``gaps'' of size $K_{\max}$ at each value location. Special BOV (beginning-of-value) tokens mark where generation should occur. During the first inference step, the model outputs next-token probabilities for the entire input, and HPD selects $N$ probabilities at the BOV positions to generate the first token of each value simultaneously. Subsequent tokens are appended to the end of the sequence but assigned position IDs that logically place them in the previously created gaps, maintaining key-value cache functionality. A position-based causal attention mask ensures tokens only attend to appropriate context despite being arranged out of order in physical memory. In subsequent steps, only the $N$ new tokens are passed to the model with the cached keys/values, generating the $t$-th token for all values in parallel until completion. HPD further stacks $J$ documents in a single prompt, decoding $J \times N$ tokens per step; combined with batch inference (batch size $b$), this yields $b \times J \times N$ tokens per step, achieving large efficiency gains.

HPD requires custom fine-tuning for parallel decoding. BOV tokens are added to the vocabulary, and a ``block ID'' tensor tracks which inference step each token belongs to (block 0 for the prompt, block 1 and above for generated tokens). Position IDs are manipulated during training to match the inference configuration, and a custom 2D attention mask enforces that prompt tokens only attend to the prompt while generated tokens attend to themselves and lower block IDs. This enables the model to match autoregressive quality while maintaining $\sim$10$\times$ speedup during inference.

\paragraph{Attribute Value Standardization}

A key challenge in large-scale attribute extraction is ensuring that extracted values are consistent and comparable across products, categories, and marketplaces. Without standardization, the same attribute can produce highly variable surface forms (e.g., ``30 oz'', ``30 ounces'', ``30oz'', ``thirty ounces'') that are difficult to use in downstream applications requiring exact matching or aggregation. We address this at multiple levels: Stage~1 defines expected data types and standardized categorical values for each attribute; Stage~2 prompts instruct the model to adhere to type requirements, output numerical values with standard units, select from predefined categorical options, and produce comma-separated lists for multi-valued attributes; and a strictly structured JSON output format further reduces variability across runs.

We additionally experiment with Constrained Hyper-Parallel Decoding at inference time, where token-level constraints force the model to generate only valid data types and standardized categorical values. For categorical attributes, a logit mask is applied at each decoding step to permit only tokens corresponding to valid options from the Stage~1 schema; for numerical attributes, a regex expression enforces the correct number-unit format (e.g., allowing only digit and unit tokens in sequence). In practice, we find that the combination of careful prompt engineering and knowledge distillation from a foundational LLM during fine-tuning already yields high standardization compliance, with constrained decoding affecting fewer than 1\% of extracted values while providing a formal guarantee of format adherence.

\section{Experiments and Results}

\subsection{Experimental Setup}
\label{sec:experimental_details}

\paragraph{Product Catalog and Attribute Schema Inventory.}
We perform experiments with our pipeline on a large proprietary e-commerce catalog containing more than 30M products spanning a few thousand product categories. Each product is represented by catalog text and metadata, including title, description, and other structured or semi-structured product details. Stage~1 produces category-level key attribute schemas, which are then standardized across categories. The resulting schema inventory contains several thousand unique attributes. Across the evaluated catalog, these attributes correspond to tens of millions of observed attribute values, including numerical, categorical, and free-form values.

\paragraph{Training data generation.}
To train the Stage~2 extraction model, we construct a supervised dataset of 100K product-schema examples sampled across product categories. For each example, we prompt the Claude Sonnet 4 LLM with the product catalog context and corresponding Stage~1 attribute schema to produce a structured JSON object containing the extracted value for each attribute, or null when absent. This data is used for HPD fine-tuning.

\paragraph{Models and HPD configuration.}
We fine-tune Qwen3-4B and Qwen3-8B models for purchase-discriminative attribute value generation using Hyper-Parallel Decoding. Each product is paired with a fixed category-level schema, and the model generates one value slot per attribute in structured JSON format. We use up to 20 attributes per product with $K_{\max}=100$ tokens and 6 input products per prompt. Fine-tuning is performed on an AWS EC2 p4de.24xlarge instance with 8 A100 GPUs.

At inference time, HPD decodes values for all attributes in parallel, rather than generating the JSON output strictly left-to-right. We compare against a foundational LLM baseline accessed via batched API inference, which processes requests asynchronously with variable queue wait times.

\paragraph{Evaluation.}
We comprehensively evaluate both stages using automated and human evaluation. For Stage~1, we use Claude Sonnet 4.5 as an LLM judge to classify discovered attributes as \emph{correct/relevant}, \emph{irrelevant}, or \emph{too vague/too specific}. For Stage~2, we use the same LLM judge to evaluate extraction quality on 20K products, classifying outputs as: \emph{correct extraction}, \emph{correct null} (attribute absent and model correctly returns null), \emph{incorrect extraction}, \emph{missed extraction} (value present but model returns null), or \emph{ungrounded extraction}. We additionally perform human evaluation on 500 products. Finally, to assess determinism, we run the complete pipeline five times on 50 products from diverse categories and compare outputs across runs.

\paragraph{Attribute Schema Discovery Quality}

We observe that 89.6\% of the discovered key attribute names are classified as ``correct'' or relevant by the LLM judge with respect to the category, 4.0\% are deemed irrelevant, and 6.4\% are considered either too vague or too specific for the given category (e.g., ``Performance'' is overly vague as it lacks discriminative specificity, while ``Bluetooth Codec Support'' is too specific for a broad ``Audio Equipment'' category). Table~\ref{tab:stage1_results} summarizes the schema quality evaluation. This assessment is conducted at the category level across a few thousand categories, covering 20 key attributes per category.

\begin{table}[t]
\centering
\small
\begin{tabular}{lr}
\toprule
\textbf{Classification} & \textbf{\% of Attributes} \\
\midrule
Correct / Relevant & 89.6\% \\
Too vague or too specific & 6.4\% \\
Irrelevant & 4.0\% \\
\bottomrule
\end{tabular}
\caption{Stage~1 attribute schema quality evaluated by an LLM judge across a few thousand product categories (20 attributes per category).}
\label{tab:stage1_results}
\end{table}

\paragraph{Attribute Value Generation Quality}

As shown in Table \ref{tab:stage2_results}, the accuracy of key attribute extraction is very similar for both fine-tuned models, at approximately 85\%. Both Qwen3 models fine-tuned with HPD achieve extraction accuracy on par with or slightly exceeding the foundational teacher LLM, demonstrating that knowledge distillation combined with task-specific fine-tuning can fully close the quality gap despite a $>$100$\times$ reduction in model parameters. Notably, HPD introduces no quality degradation while providing significant speed gains.

Of the remaining approximately 15\%, the LLM judge flags 4.6\% as \textit{potentially ungrounded} extractions (values not directly verifiable from the explicit input text), 3.5\% as missed extractions, and 2\% as incorrect extractions. Manual investigation of 50 such cases, corroborated by our 500-product human evaluation, reveals that approximately half of ungrounded cases are reasonable inferences from implicit context (e.g., inferring ``plastic'' from ``BPA-free''), bringing the effective ungrounded extraction rate to approximately 2.3\%. Additionally, fewer than 2\% involve format violations where the model does not conform to Stage~1 type constraints. The remaining 3-4\% include ambiguous cases due to catalog noise, where the LLM judge cannot make a definitive determination, and are excluded from evaluation.

\begin{table*}[t]
\centering
\small
\setlength{\tabcolsep}{6pt}
\begin{tabular}{lrrrrr}
\toprule
\textbf{Model}& \textbf{\makecell{Total\\Correct}}
  & \textbf{\makecell{Correct\\Extractions}}
  & \textbf{\makecell{Correct Null\\Extractions}}
  & \textbf{\makecell{Incorrect\\Extractions}}
  & \textbf{\makecell{Missed\\Extractions}}\\
\midrule
Foundational LLM Baseline & 84.80\% & 54.30\% & 30.50\% & 1.30\% & 4.00\% \\
Qwen3-4B (AR) & 84.60\% & 51.80\% & 32.80\% & 2.00\% & 3.80\% \\
Qwen3-4B (HPD) & \textbf{85.40\%} & \textbf{52.40\%} & 33.00\%          & \textbf{1.70\%} & \textbf{3.50\%} \\
Qwen3-8B (HPD) & 84.40\%          & 49.60\%          & \textbf{34.80\%} & 2.60\%          & \textbf{3.50\%} \\
\bottomrule
\end{tabular}
\caption{Stage~2 key attribute value generation quality for fine-tuned models, evaluated
by an LLM judge (Claude Sonnet 4.5) on a sample of 20K products. Bold indicates the
best result per column (highest for correct metrics, lowest for incorrect/missed).}
\label{tab:stage2_results}
\end{table*}

Interestingly, the larger Qwen3-8B model performs slightly worse than Qwen3-4B. Our practical constraints - processing several hundred million products with daily incremental and monthly full catalog refreshes - require sustained high throughput on commodity GPU instances, where models exceeding 8B parameters significantly reduce per-GPU batch capacity. We therefore focus on models at or below 8B parameters and rely on knowledge distillation to close any quality gap. %

\paragraph{Determinism and Coverage.} To assess reliability, we run both stages five times on 50 products from diverse categories. For both schema discovery and value extraction, approximately 93\% of outputs are identical or semantically equivalent across runs, with variations primarily in subjective attributes such as ``special features.'' Overall pipeline coverage is 99.997\%: attributes are successfully generated for all products except those lacking catalog information or with incorrect category assignments (0.003\% of products).

\paragraph{Human Evaluation}
We perform human evaluation on 500 products spanning diverse categories, assessing correctness of extracted values against input catalog text for both categorical and numerical attributes. Human judges rate 85\% of extractions as correct, consistent with the LLM judge evaluation, with Qwen3-4B again performing slightly better than Qwen3-8B. Overall, the LLM judge and human annotators agree on 91\% of generated attribute values. Of the errors, approximately 6\% are incorrect extractions; the remainder are primarily data type violations and, for numerical quantity attributes, cases where the attribute information is present only in product images rather than the catalog text provided as input.

\paragraph{Efficiency and Throughput}

Table~\ref{tab:efficiency} compares the cost and throughput of our HPD-based extraction against both standard autoregressive decoding on the same model and the foundational LLM baseline using cloud-hosted batched API inference. HPD achieves an average throughput of $\sim$159K products per hour per instance. Running inference on 30M products using 5 parallel instances takes approximately 36 hours at a total cost of $\sim$\$5,000. Standard autoregressive decoding on the same Qwen3-4B model yields approximately 18K products per hour per instance ($\sim$9$\times$ lower throughput), %
requiring over 330 hours on the same hardware at a cost of $\sim$\$46K.

The foundational LLM baseline, accessed via cloud-hosted batched API inference with 10 parallel batch jobs, achieves a nominal throughput of $\sim$100K products per hour excluding queue wait times. In practice, variable queue delays add significant additional latency, making the effective wall-clock time substantially longer than the reported 300 hours. The total API cost for 30M products is $\sim$\$63,000, representing a 92\% cost reduction with HPD. Fine-tuning the Qwen3-4B model requires 8 hours on a single instance with 8 A100 GPUs, at a one-time cost of $\sim$\$350.

\paragraph{Effect of Attribute Schema Size on Parallelism.}

HPD decodes all $N$ attribute values simultaneously, generating $J \times N$ tokens per inference step. Since decoding steps remain fixed at $K_{\max}$ regardless of $N$, throughput in attribute-values per hour scales approximately linearly with schema size. We validate this by varying $N \in \{10, 12, 16, 18, 20\}$ and observe that product throughput remains approximately constant while attribute-values extracted per hour increases proportionally. Meanwhile, the foundational LLM baseline cost scales linearly with $N$, yielding progressively larger cost reductions from approximately 84\% at $N=10$ to 92\% at $N=20$, making our choice of $N=20$ near-optimal for both parallelism and category coverage.

\begin{table}[t]
\centering
\small
\resizebox{\columnwidth}{!}{%
\begin{tabular}{l r r r}
\toprule
& \makecell[r]{\textbf{Qwen3-4B}\\(HPD)} & \makecell[r]{\textbf{Qwen3-4B}\\(AR)} & \makecell[r]{\textbf{Foundational}\\\textbf{LLM}} \\
\midrule
Throughput (products/hr) & 159K & 18K & 100K$^\dagger$ \\
Parallel instances / jobs & 5 & 5 & 10 \\
Time for 30M products & 36 hrs & 333 hrs & 300+ hrs$^\dagger$ \\
Inference cost & \$5K & \$46K & \$63K \\
\makecell[l]{Cost reduction vs.\\Foundational LLM} & \textbf{92\%} & 27\% & --- \\
\bottomrule
\multicolumn{4}{l}{\scriptsize $^\dagger$ Excludes variable queue wait times, which add significant latency.}
\end{tabular}%
}
\caption{Efficiency comparison for 30M products between HPD inference, standard autoregressive (AR) inference, and foundational LLM batched API inference.}
\label{tab:efficiency}
\end{table}

\section{Practical Use Case Evaluation}

We employ our full two-stage pipeline to process product catalogs across two major marketplaces, extracting key attributes for hundreds of millions of products.
Large-scale inference is distributed across 4--20 A100 GPU instances, each processing products independently to achieve linear throughput scaling. The pipeline is containerized for reproducibility across instance types.

Given that new products appear in the catalog daily and product content is frequently updated, we plan a tiered refresh strategy. New products are processed on a daily or weekly cadence, while the full catalog is re-processed monthly to capture updates. Schema discovery (Stage~1) is refreshed less frequently as category-level attributes are relatively stable; schemas are versioned per refresh cycle and validated against the previous version before deployment. For Stage~2, outputs that fail JSON parsing or violate data type constraints are caught by automated post-processing validation and re-processed in the subsequent batch.

The extracted attribute-value pairs are published to an internal large-scale knowledge base and made queryable by downstream use cases. These structured representations can enable several applications: (i) product comparison via shared category-level schemas; (ii) catalog enrichment, where extracted values fill gaps in incomplete product listings and augment sparse catalog metadata; (iii) catalog quality improvement by surfacing inconsistencies in product details; and (iv) personalized product understanding through compact structured inputs to recommendation and search systems.

We additionally validate our outputs through live seller interviews across two marketplaces, where sellers evaluate the extracted attributes for quality and usefulness in differentiating products and informing customer purchase decisions. Sellers provide positive feedback, confirming that the extracted attributes reflect factors they consider when positioning their products relative to competitors.

\subsection{Sample System Output}
\label{sec:sample_output}

\begin{table}[ht]
\centering
\small
\resizebox{\columnwidth}{!}{%
\begin{tabular}{lll}
\toprule
\textbf{Attribute} & \textbf{Data Type} & \textbf{Standardized Values} \\
\midrule
Brew Type & categorical & \makecell[l]{Manual, Semi-Automatic,\\Automatic, Super-Automatic, Pod} \\
Pump Pressure & numerical & (bars) \\
Heating System & categorical & \makecell[l]{Thermocoil, Single Boiler,\\Dual Boiler, Thermoblock} \\
Temperature Control & categorical & PID, Thermostat, Digital \\
Grinder Type & categorical & Conical Burr, Flat Burr, Blade, None \\
Tank Capacity & numerical & (oz) \\
Milk System & categorical & Steam Wand, Auto Frother, None \\
Build Material & categorical & Stainless Steel, Plastic, Aluminum \\
\bottomrule
\end{tabular}%
}
\caption{Stage~1 output: discovered attribute schema for \texttt{COFFEE\_MAKER}.}
\label{tab:sample_schema}
\end{table}

We illustrate the end-to-end pipeline output for the product category \texttt{COFFEE\_MAKER}. Stage~1 discovers a schema of purchase-discriminative attributes, which we show in \Cref{tab:sample_schema}.
Given this schema and the catalog text of a specific product (title, description, and bullet points), Stage~2 extracts the following structured attribute-value pairs:\\
{\small
\noindent\begin{minipage}{\columnwidth}
\begin{verbatim}
{"Brew Type": "Semi-Automatic",
 "Pump Pressure": "15 bars",
 "Heating System": "Thermocoil",
 "Temperature Control": "PID",
 "Grinder Type": "Conical Burr",
 "Tank Capacity": "67 oz",
 "Milk System": "Steam Wand",
 "Build Material": "Stainless Steel"}
\end{verbatim}
\end{minipage}
}

All products in the \texttt{COFFEE\_MAKER} category share this same schema, making their structured representations directly comparable. For example, a budget pod machine would yield \texttt{\{``Brew Type'': ``Pod'', ``Heating System'': ``Thermoblock'', ``Grinder Type'': ``None'', ``Build Material'': ``Plastic'', ...\}}, enabling direct attribute-level comparison.

\section{Conclusion}

We presented a scalable two-stage pipeline for extracting purchase-discriminative attributes from large e-commerce catalogs. The first stage automatically discovers compact, category-level schemas using a foundational LLM, eliminating manual schema engineering across thousands of categories. The second stage extracts attribute values at scale using a fine-tuned Qwen3-4B model with HPD, which exploits the conditional independence of attribute values to decode them simultaneously. The system achieves 85\% extraction accuracy, on par with the foundational teacher LLM, while reducing inference costs by 92\%. Future directions include incorporating multimodal inputs such as product images to capture visually conveyed attributes, and extending the pipeline to multilingual catalogs. %

\section*{Limitations}

Our system extracts attributes exclusively from textual catalog content such as product titles, descriptions, and bullet points. Product information that is conveyed only through images, such as visual design details, color variations, or quantities shown in packaging photos, is not captured by our current approach. Incorporating multimodal inputs to address this gap is left for future work. The current pipeline has been evaluated only on English-language catalogs. Extending the system to non-English marketplaces would require multilingual fine-tuning and potentially different schema discovery prompts, which we have not yet explored. The conditional independence assumption underlying HPD assumes that extracting the value of one attribute should not depend on previously extracted values. While this holds for most attributes in practice, certain attributes may exhibit correlations (e.g., product weight and volume, or material and durability). Our empirical results suggest this does not significantly impact extraction quality, but a deeper investigation into correlated attribute groups is warranted. Finally, we evaluated our fine-tuned models exclusively using the Qwen3 model family in 4B and 8B parameter sizes. Other model architectures or larger model sizes may yield different quality-efficiency tradeoffs, and the optimal model choice may vary across domains or catalog characteristics.

\bibliography{custom}

@inproceedings{yang2022mave,
  title = {MAVE: A Product Dataset for Multi-source Attribute Value Extraction},
  author = {Yang, Li and Wang, Qifan and Yu, Zac and Kulkarni, Anand and Sanghai, Sumit Kumar and Shu, Bin and Elsas, Jon and Kanagal, Bhargav},
  booktitle = {Proceedings of the Fifteenth ACM International Conference on Web Search and Data Mining},
  pages = {1256--1265},
  year = {2022},
  doi = {10.1145/3488560.3498377}
}

@article{brinkmann2023extractgpt,
  title = {ExtractGPT: Exploring the Potential of Large Language Models for Product Attribute Value Extraction},
  author = {Brinkmann, Alexander and Shraga, Roee and Bizer, Christian},
  journal = {arXiv preprint arXiv:2310.12537},
  year = {2023}
}

@inproceedings{khandelwal2023large,
  title = {Large Scale Generative Multimodal Attribute Extraction for E-commerce Attributes},
  author = {Khandelwal, Anant and Mittal, Happy and Kulkarni, Shreyas and Gupta, Deepak},
  booktitle = {Proceedings of the 61st Annual Meeting of the Association for Computational Linguistics (Volume 5: Industry Track)},
  pages = {305--312},
  year = {2023},
  address = {Toronto, Canada},
  publisher = {Association for Computational Linguistics},
  doi = {10.18653/v1/2023.acl-industry.29}
}

@inproceedings{blume2023generative,
  title = {Generative Models for Product Attribute Extraction},
  author = {Blume, Ansel and Zalmout, Nasser and Ji, Heng and Li, Xian},
  booktitle = {Proceedings of the 2023 Conference on Empirical Methods in Natural Language Processing: Industry Track},
  pages = {575--585},
  year = {2023},
  address = {Singapore},
  publisher = {Association for Computational Linguistics},
  doi = {10.18653/v1/2023.emnlp-industry.55}
}

@inproceedings{shinzato2023unified,
  title = {A Unified Generative Approach to Product Attribute-Value Identification},
  author = {Shinzato, Keiji and Yoshinaga, Naoki and Xia, Yandi and Chen, Wei-Te},
  booktitle = {Findings of the Association for Computational Linguistics: ACL 2023},
  pages = {6599--6612},
  year = {2023},
  address = {Toronto, Canada},
  publisher = {Association for Computational Linguistics},
  doi = {10.18653/v1/2023.findings-acl.413}
}

@inproceedings{xu2023towards,
  title = {Towards Open-World Product Attribute Mining: A Lightly-Supervised Approach},
  author = {Xu, Liyan and Zhang, Chenwei and Li, Xian and Shang, Jingbo and Choi, Jinho D.},
  booktitle = {Proceedings of the 61st Annual Meeting of the Association for Computational Linguistics (Volume 1: Long Papers)},
  pages = {12223--12239},
  year = {2023},
  address = {Toronto, Canada},
  publisher = {Association for Computational Linguistics},
  doi = {10.18653/v1/2023.acl-long.683}
}

@inproceedings{chen2023named,
  title = {Does Named Entity Recognition Truly Not Scale Up to Real-world Product Attribute Extraction?},
  author = {Chen, Wei-Te and Shinzato, Keiji and Yoshinaga, Naoki and Xia, Yandi},
  booktitle = {Proceedings of the 2023 Conference on Empirical Methods in Natural Language Processing: Industry Track},
  pages = {152--159},
  year = {2023},
  address = {Singapore},
  publisher = {Association for Computational Linguistics},
  doi = {10.18653/v1/2023.emnlp-industry.16}
}

@inproceedings{yang2024eave,
  title = {EAVE: Efficient Product Attribute Value Extraction via Lightweight Sparse-layer Interaction},
  author = {Yang, Li and Wang, Qifan and Chi, Jianfeng and Liu, Jiahao and Wang, Jingang and Feng, Fuli and Xu, Zenglin and Fang, Yi and Huang, Lifu and Liu, Dongfang},
  booktitle = {Findings of the Association for Computational Linguistics: EMNLP 2024},
  pages = {1491--1505},
  year = {2024},
  address = {Miami, Florida, USA},
  publisher = {Association for Computational Linguistics},
  doi = {10.18653/v1/2024.findings-emnlp.80}
}

@inproceedings{zhang2024stronger,
  title = {{Stronger, Lighter, Better: Towards Life-Long Attribute Value Extraction for E-Commerce Products}},
  author = {Zhang, Tao and Zhang, Chenwei and Li, Xian and Shang, Jingbo and Nguyen, Hoang and Yu, Philip},
  booktitle = {Findings of the Association for Computational Linguistics: ACL 2024},
  pages = {8631--8643},
  year = {2024},
  address = {Bangkok, Thailand},
  publisher = {Association for Computational Linguistics},
  doi = {10.18653/v1/2024.findings-acl.510}
}

@inproceedings{glavas2026breaking,
  title={Breaking the Autoregressive Chain: Hyper-Parallel Decoding for Efficient LLM-Based Attribute Value Extraction},
  author={Glavas, Theodore and Vedula, Nikhita and Dhyani, Dushyanta and Zhu, Yilun and Malmasi, Shervin},
  booktitle={Findings of the Association for Computational Linguistics: ACL 2026},
  pages={36792--36808},
  year={2026}
}

@article{huang2025attributeforge,
  title = {{AttributeForge: An Agentic LLM Framework for Automated Product Schema Modeling}},
  author = {Huang, Yunhan and Ramo, Klevis and Iovine, Andrea and Monteiro, Melvin and Gokalp, Sedat and Bakshi, Arjun and Turalic, Hasan and Kumar, Arsh and Neumeier, Jona and Yates, Ripley and Monir, Rejaul and Hartmann, Simon and Manglik, Tushar and Yakout, Mohamed},
  year = {2025}
}

@inproceedings{zheng2018opentag,
  title = {OpenTag: Open Attribute Value Extraction from Product Profiles},
  author = {Zheng, Guineng and Mukherjee, Subhabrata and Dong, Xin Luna and Li, Feifei},
  booktitle = {Proceedings of the 24th ACM SIGKDD International Conference on Knowledge Discovery \& Data Mining},
  pages = {1049--1058},
  year = {2018},
  doi = {10.1145/3219819.3219839}
}

@inproceedings{wang2020learning,
  title = {Learning to Extract Attribute Value from Product via Question Answering: A Multi-task Approach},
  author = {Wang, Qifan and Yang, Li and Kanagal, Bhargav and Sanghai, Sumit and Sivakumar, D. and Shu, Bin and Yu, Zac and Elsas, Jon},
  booktitle = {Proceedings of the 26th ACM SIGKDD International Conference on Knowledge Discovery \& Data Mining},
  year = {2020}
}

@article{weikum2021machine,
  title={Machine Knowledge: Creation and Curation of Comprehensive Knowledge Bases},
  author={Weikum, Gerhard and Dong, Xin Luna and Razniewski, Simon and Suchanek, Fabian},
  journal={Foundations and Trends in Databases},
  year={2021}
}

@article{bian2025llmkg,
  title={LLM-empowered Knowledge Graph Construction: A Survey},
  author={Bian, Haonan},
  journal={arXiv preprint arXiv:2510.20345},
  year={2025}
}

@inproceedings{hongwimol2026autopkg,
  title={Auto{PKG}: An Automated Framework for Dynamic E-commerce Product-Attribute Knowledge Graph Construction},
  author={Hongwimol, Pollawat and Shang, Haoning and Wang, Chutong and Wan, Zhichao and Gao, Yi and Li, Yuanming and Gui, Lin and Sun, Wenhao and Yu, Cheng},
  booktitle={Findings of ACL},
  year={2026}
}

@inproceedings{peshevski2025agent,
  title={AI Agent-Driven Framework for Automated Product Knowledge Graph Construction in E-Commerce},
  author={Peshevski, Dimitar and Stojanov, Riste and Trajanov, Dimitar},
  booktitle={Proceedings of the 1st GOBLIN Workshop on Knowledge Graph Technologies},
  year={2025}
}

@misc{glavas2026intrapromptparalleldecodingcommoncontext,
      title={Intra-Prompt Parallel Decoding for Common-Context Question Answering}, 
      author={Theodore Glavas and Nikhita Vedula and Dushyanta Dhyani and Antonios Valkanas and Yilun Zhu and Shervin Malmasi},
      year={2026},
      eprint={2609.05707},
      archivePrefix={arXiv},
      primaryClass={cs.CL},
      url={https://arxiv.org/abs/2609.05707}, 
}

\appendix

\end{document}